\documentclass[letterpaper, 10 pt, conference]{ieeeconf}  

\IEEEoverridecommandlockouts                              

\usepackage{graphics} 
\usepackage{mathptmx} 
\usepackage{times} 
\usepackage{amsmath} 
\usepackage{amssymb}  

\usepackage{comment}
\usepackage{todonotes}
\usepackage{url}
\usepackage{booktabs}
\usepackage{capt-of}
\usepackage{graphicx}
\usepackage{subcaption}
\usepackage{pgfplots}
\usepackage{tikz}
\usepackage{nicefrac}
\usepackage{algorithm}
\usepackage{algpseudocode}
\let\labelindent\relax
\usepackage{enumitem}
\renewcommand{\textcolor}[2]{#2}

\pgfmathsetseed{\number\pdfrandomseed} 

\title{\LARGE \bf
Evaluating UAV Path Planning Algorithms for Realistic Maritime Search and Rescue Missions
}

\author{Martin Messmer$^{1}$ and Andreas Zell$^{1}$
\thanks{*This work was supported by the German Ministry for Economic Affairs and Climate Action under grant number FKZ: 19A21009C}
\thanks{$^{1}$Faculty of Computer Science, University of Tuebingen, 72076 Tübingen, Germany
        {\tt\small martin.messmer@uni-tuebingen.de}}%
}

\begin{document}

\maketitle
\thispagestyle{empty}
\pagestyle{empty}

\begin{abstract}

Unmanned Aerial Vehicles (UAVs) are emerging as very important tools in search and rescue (SAR) missions at sea, enabling swift and efficient deployment for locating individuals or vessels in distress. The successful execution of these critical missions heavily relies on effective path planning algorithms that navigate UAVs through complex maritime environments while considering dynamic factors such as water currents and wind flow. Furthermore, they need to account for the uncertainty in search target locations. However, existing path planning methods often fail to address the inherent uncertainty associated with the precise location of search targets and the uncertainty of oceanic forces. In this paper, we develop a framework to develop and investigate trajectory planning algorithms for maritime SAR scenarios employing UAVs. We adopt it to compare multiple planning strategies, some of them used in practical applications by the United States Coast Guard. Furthermore, we propose a novel planner that aims at bridging the gap between computation heavy, precise algorithms and lightweight strategies applicable to real-world scenarios.
\end{abstract}

\section{Introduction}

The increasing adoption of Unmanned Aerial Vehicles (UAVs) for maritime search and rescue (SAR) missions has introduced novel operational capacities. Specifically, UAVs provide enhanced endurance and real-time data transmission, which can complement traditional human-led SAR efforts. Nonetheless, devising efficient path planning for UAVs in this context remains challenging, primarily due to the variable and unpredictable characteristics of the maritime environment.

\begin{figure}[ht]
        \centering\includegraphics[width=0.49\textwidth]{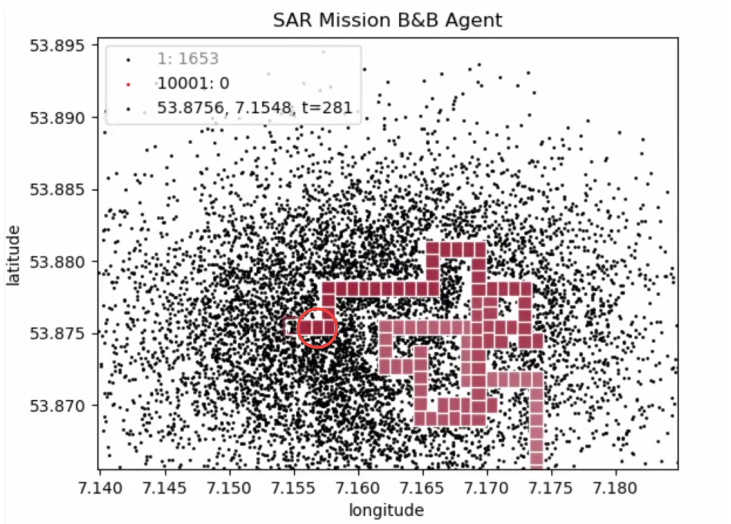}
    \caption{Example of a trajectory of the branch and bound agent right after it found a search target. Here, it succeeded in doing so after approximately 30 minutes after take-off. The search target's position is highlighted by a circle. The plot is taken from our framework.}
    \label{fig:bnb_example}
    \vspace{-5mm}
\end{figure}

In maritime SAR scenarios, efficient coverage of the search area and accurate target localization are pivotal. Many popular existing path planning algorithms, like A* or Dijkstra~\cite{russell2010artificial}, operate under the assumption of known and static environmental conditions. These assumptions are not applicable to maritime SAR for multiple reasons: Factors such as water currents and wind dynamics determine the search targets' trajectory, affecting its position and velocity. However, these factors are known very imprecisely at best. Furthermore, imprecise knowledge about locations of search targets in distress calls for probabilistic trajectory planning algorithms.

Extensive research in the area of maritime search and rescue utilizing Unmanned Aerial Vehicles has predominantly centered on computer vision, with a specific emphasis on object detection. This focus resulted in the publication of dedicated datasets \cite{varga2022seadronessee}, works that delve into the intricacies of detecting small objects \cite{lee2018snip, varga2021tackling}, exploration into the integration of supplemental sensor data to enhance detection efficacy \cite{messmer2022gaining, kiefer2022leveraging}, and contributions from control theory \cite{raap2019moving}.

Therefore, there is an evident need for a more adaptive approach in UAV path planning for maritime SAR missions -- one that considers both target location uncertainty and dynamic environmental factors. Hence, we analyze competitive path planning algorithms. All of them integrate probabilistic models to account for target location uncertainty and use real-time meteorological data to adapt to changing water currents and wind conditions.

Furthermore, this paper introduces a novel path planning algorithm, building upon the foundational principles of branch-and-bound (BnB) techniques \cite{sato2008path}. Our proposed method aims to bridge the gap from theory to practical application by leveraging the strengths of existing theory on BnB algorithms while tailoring them for real-world application. It takes into account environmental factors, such as water current and wind flow. Yet, in contrast to the existing literature on BnB-based path planning \cite{sato2008path, raap2017trajectory, raap2019moving}, our approach is designed to be computationally lightweight.

The main contributions of this paper are the following:

\begin{itemize}
    \item We propose a novel trajectory planning algorithm that aims at bridging the gap from easily applicable algorithms that take almost no environmental data into account to computation-heavy and theory-backed algorithms like branch-and-bound algorithms.
    \item The algorithms in this paper were evaluated using a newly developed framework for researching and testing maritime SAR algorithms for UAVs. It is available on GitHub\footnote{\url{https://github.com/cogsys-tuebingen/pathplanningrepository}}. It is fully written in Python, which makes it easy to use.
    \item We compare and evaluate multiple trajectory planning algorithms and discuss their results.
\end{itemize}

The structure of this paper is as follows: Section 2 reviews the current research in this area. Section 3 explains the algorithms and methods used in this study, including those related to the change of search targets over time and the main path planning algorithms. Section 4 presents the experimental results and a subsequent discussion. Finally, section 5 discusses possible future research and the limitations of this paper.
\vspace{-3mm}

\section{Related Work}

In \cite{martinez2021search}, the authors build a pipeline to perform search operations from a UAV equipped with a smartphone. They also record a dataset for the training of their neural network. While this is interesting work towards UAV-based SAR missions, they do not investigate the path planning problem. Similarly, there is a vast number of publications on computer vision from UAVs \cite{varga2022seadronessee, varga2021tackling, messmer2022gaining, kiefer2022leveraging, du2019visdrone}, which is very important for automated SAR scenarios. However, it doesn't address the problem of how to compute the UAV's trajectory.
The authors in \cite{sato2008path} construct bounds for a branch and bound algorithm for the search problem with a single UAV. This finds an optimal solution to the problem. However, the algorithm works, as shown, merely on problems in the range of $10\times 10$ to $20 \times 20$ grids, which is far too small for any realistic application. While \cite{morin2010ant} uses an ant-colony optimization method, it suffers from the same problem. 
In \cite{berger2013exact} the authors propose a mixed integer linear programming approach to solve this problem. While this also delivers exact solutions, it is again computationally too intensive for application in practice.
In \cite{riehl2007cooperative}, the authors use graph-based model-predictive search to solve problems in the range of roughly $34 \times 34$ grids. That is already larger but still too small for most real-world applications. For comparison, in our experiments we usually used a grid size of $2500 \times 2500$, see section \ref{sec:experiments}.
The authors in \cite{dagestad2018opendrift} developed OpenDrift a framework to efficiently simulate drift of objects or substances in the ocean, such as oil spills, floating debris, life-rafts or vessels in distress. We will employ this work for the latter. 
Similarly, \cite{wu2023modeling} model the leeway drift of people in water. While this is very important, they don't investigate trajectory planning for maritime SAR missions. In \cite{guoxiang2010sargis} the authors describe a full application to plan maritime SAR missions. They use trajectory planning methods similar to the spiral and boustrophedon search used later in this work. \\
In \cite{kratzke2010search}, the authors describe the planner model used by the United States Coast Guard; their environmental model is basically equal to the method of OpenDrift. How the planning algorithm works, however, is not disclosed in detail. \\
The authors of \cite{li2023survey} investigate the sensor, communication, and control subsystems of a UAV platform potentially employed for maritime SAR missions. This is highly relevant for SAR missions, yet it gives no insight into path planning methods for the problem at hand. The work \cite{tiemann2018supporting} investigates the problem where the search target in distress is continuously sending a distress signal and use this to enhance the estimated target position. While distressed ships might be able to provide such a search aid, live rafts may not. Hence we explore the case, where the search target is not transmitting signals.

\section{Method}
\label{sec:methods}

\begin{figure}
    \includegraphics[width=0.49\textwidth]{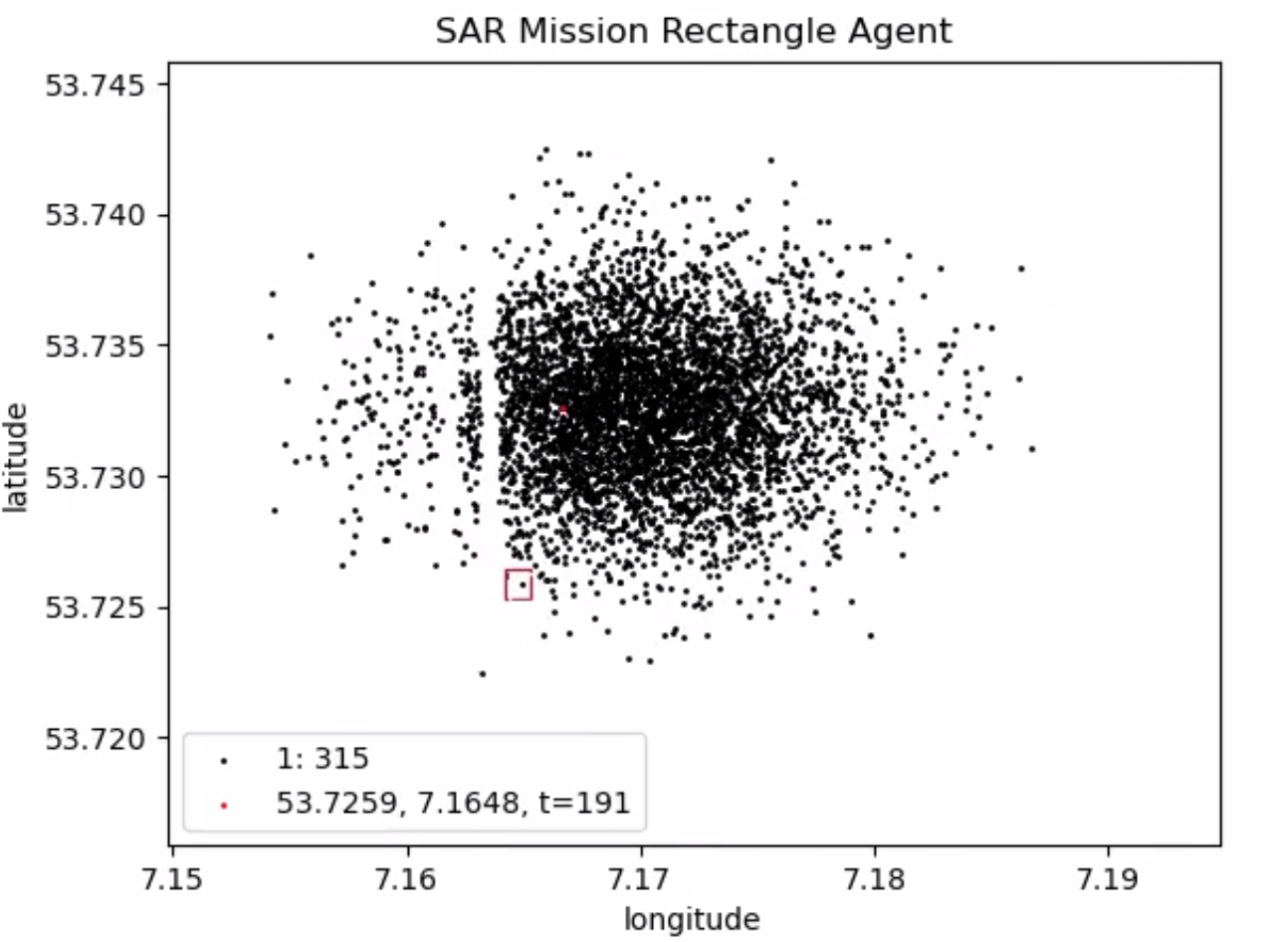}
    \includegraphics[width=0.49\textwidth]{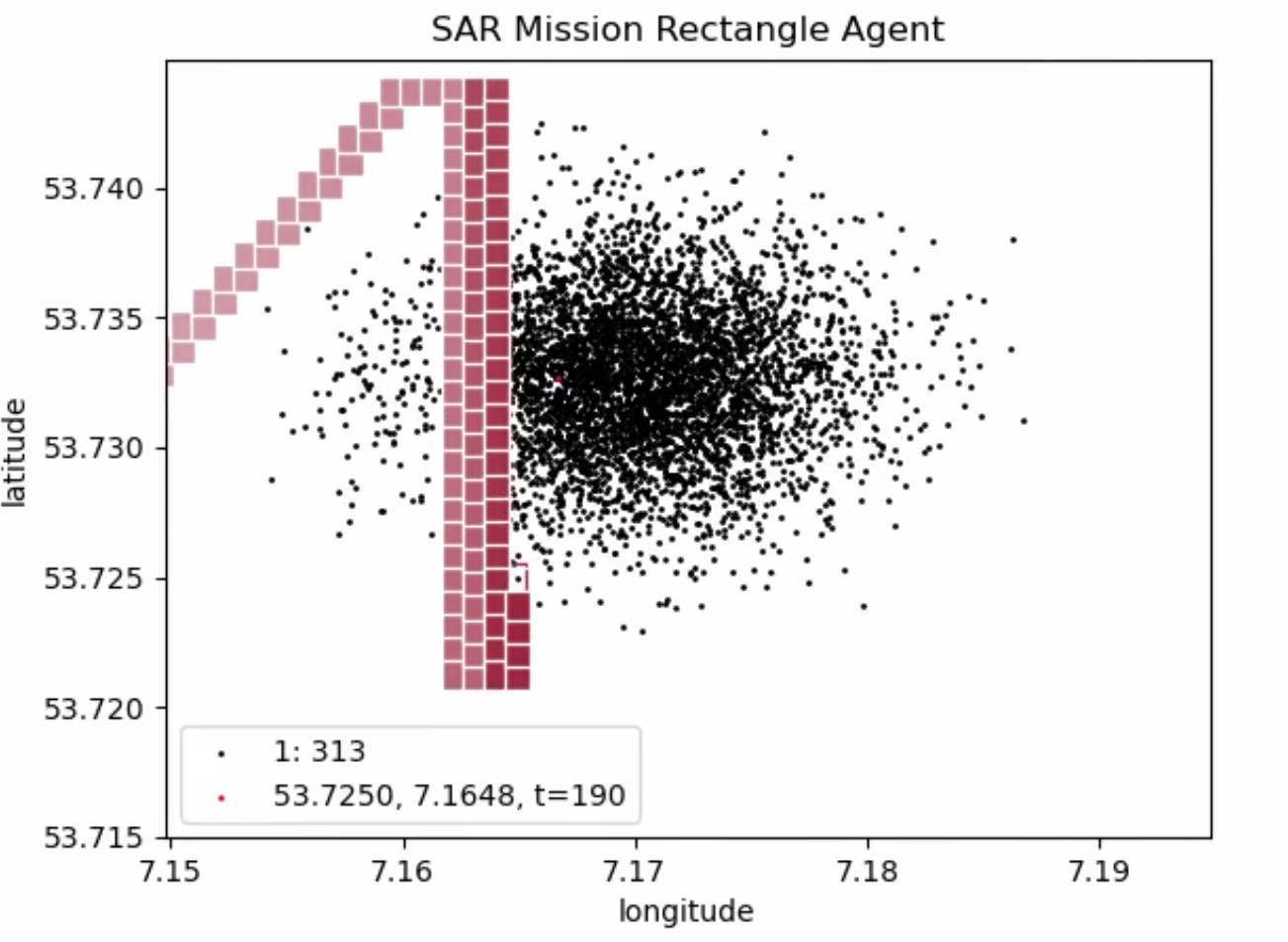}
    \caption{Two example plots taken from our framework. The agent performs the boustrophedon rectangle method. At the bottom, the agent is plotted with the trace of its trajectory for better overview of its performance. Recently visited grid cells are plotted in dark red while cells which were visited longer ago are brighter. At the top, the plot contains no trace to have a better look at the particles. The legends contain position, time, and found particles.}
    \label{fig:example_plot}
    \vspace{-5mm}
\end{figure}

\begin{figure}[ht!]
	\centering\includegraphics[width=0.85\linewidth]{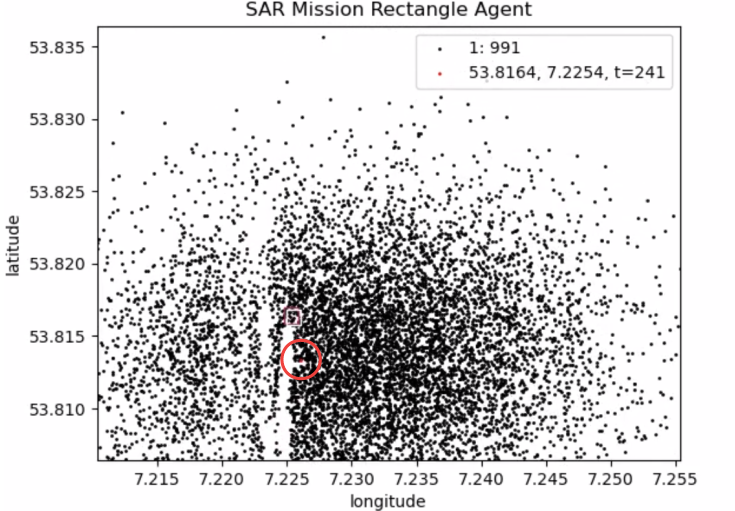} \\
	\centering\includegraphics[width=0.85\linewidth]{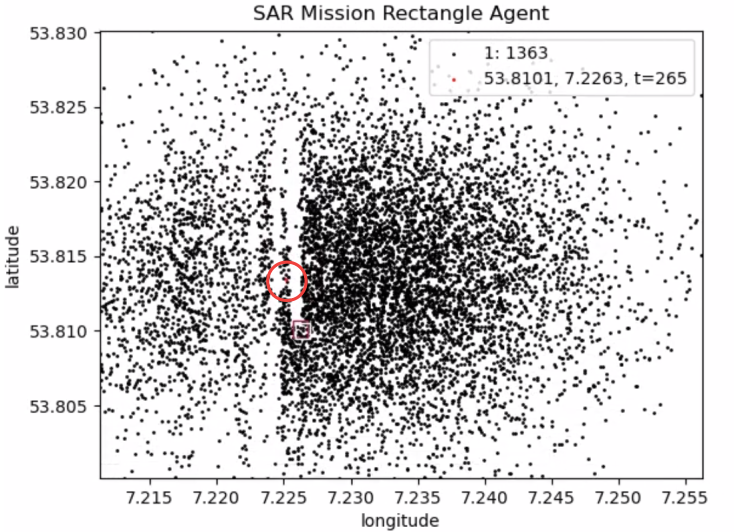}
	\caption{An unfavorable case for boustrophedon search: The two images show two closely consecutive moments in a target search, as can be seen by the simulation time in the top right corner of either image. In the top image, the UAV (red and white square) is moving north in a straight line, then turning at the northern end of the search area to fly south afterwards -- this is shown in the bottom image. The search target (red particle, highlighted by a circle around it) moves west of the UAV's position while the drone is turning around at the northern edge of its rectangular search pattern. Plots are taken from our framework.}
	\label{fig:rectangle_failing}
\end{figure}

To run and test the planning algorithms at hand, we first developed a framework to simulate the flight trajectory of an UAV. It is available on GitHub${}^1$. Plots produces by our framwork are shown in Figures \ref{fig:bnb_example}, \ref{fig:example_plot}, and \ref{fig:rectangle_failing}.  Briefly summarized, it contains the following features:

\begin{itemize}[wide=0pt]
    \item The first step of every simulation in our framework is to run an OpenDrift \cite{dagestad2018opendrift} simulation. OpenDrift (Fig. \ref{fig:opendrift}, section \ref{sec:opendrift}) is an open-source framework that models drift trajectories of objects or substances in the ocean, such as oil spills, floating debris, or in our case, targets for search and rescue (modeled as life-rafts). By leveraging the capabilities of OpenDrift, our framework distributes particles in the simulated maritime environment. These particles are a non-parametric model of the probability distribution describing the potential location of the search target as this is usually not precisely known for maritime SAR missions. The distribution location of these particles is defined by the user to model the specifics of the search scenario.
    \item Subsequent to the particle distribution, our framework creates a grid in the search area. This grid serves as a defined movement space for the UAVs. The dimensions of the grid as well as the grid tile's size are freely specified by the user. Depending on the sensor, larger or smaller grid tiles might be adequate for the scenario.
    \item The main component of our framework is its ability to constantly monitor and update the state of each particle within the simulation. Once a particle is observed, our framework deletes it, see section \ref{sec:particlefilter}. That is a key feature as observed particles need not be taken into account by the UAV for subsequent planning of the trajectory.
\end{itemize}

\begin{figure*}[h]
    \centering
    \begin{subfigure}{0.3\linewidth}

\resizebox{0.99\linewidth}{.99\linewidth}{
        \begin{tikzpicture}

            \def\gridsize{10}
            \def\spiralsize{4}
            
            \foreach \i in {0,...,\gridsize} {
                \draw[gray!30, very thin] (0,\i) -- (\gridsize,\i);
                \draw[gray!30, very thin] (\i,0) -- (\i,\gridsize);
            }

            \foreach \i in {1,...,1000} {
                \pgfmathsetmacro{\x}{5 + (rand + rand + rand + rand + rand + rand + rand + rand + rand)}
                \pgfmathsetmacro{\y}{5 + (rand + rand + rand + rand + rand + rand + rand + rand + rand)}
                \ifdim \x pt>3pt
                    \ifdim \x pt<8pt
                        \ifdim \y pt>4pt
                            \ifdim \y pt<8pt
                                \fill[gray!25] (\x,\y) circle (0.05);
                            \fi
                        \fi
                    \fi
                \fi
                
                \ifdim \x pt>0pt
                    \ifdim \x pt<10pt
                        \ifdim \y pt>0pt
                            \ifdim \y pt<10pt
                                \ifdim \x pt>8pt
                                    \fill[gray!85] (\x,\y) circle (0.05);
                                \fi
                                \ifdim \x pt<3pt
                                    \fill[gray!85] (\x,\y) circle (0.05);
                                \fi
                                \ifdim \y pt>8pt
                                    \fill[gray!85] (\x,\y) circle (0.05);
                                \fi
                                \ifdim \y pt<4pt
                                    \fill[gray!85] (\x,\y) circle (0.05);
                                \fi
                            \fi
                        \fi
                    \fi
                \fi
            }

            \tikzstyle{spiral}=[->, >=stealth, shorten >=1pt]

            \draw[spiral] (5.5,5.5) -- 
            ++(1.0,0);
            \draw[spiral] (6.5,5.5) -- 
            ++(0,1.0);
            \draw[spiral] (6.5,6.5) -- 
            ++(-1.0,0) -- 
            ++(-1,0);
            \draw[spiral] (4.5,6.5) -- 
            ++(0,-1.0) -- 
            ++(0,-1);
            \draw[spiral] (4.5,4.5) -- 
            ++(1.0,0) -- 
            ++(1,0) -- 
            ++(1,0);
            \draw[spiral] (7.5,4.5) -- 
            ++(0,1.0) -- 
            ++(0,1) -- 
            ++(0,1);
            \draw[spiral] (7.5,7.5) -- 
            ++(-1.0,0) -- 
            ++(-1,0) -- 
            ++(-1,0) -- 
            ++(-1,0);
            \draw[spiral] (3.5,7.5) -- 
            ++(0,-1.0) -- 
            ++(0,-1) -- 
            ++(0,-1);
            
            \node at (3.5,4.5) {\tiny UAV};
        
        \end{tikzpicture}
}

        \caption{Example of an expanding spiral path.}
    \end{subfigure}\hfill
    \begin{subfigure}{0.3\linewidth}
\resizebox{0.99\linewidth}{.99\linewidth}{
    \begin{tikzpicture}

        \def\gridsize{10}
        
        \foreach \i in {1,...,1000} {
            \pgfmathsetmacro{\x}{5 + (rand + rand + rand + rand + rand + rand + rand + rand + rand)}
            \pgfmathsetmacro{\y}{5 + (rand + rand + rand + rand + rand + rand + rand + rand + rand)}
            \ifdim \x pt>2pt
                \ifdim \x pt<8pt
                    \ifdim \y pt>4pt
                        \ifdim \y pt<8pt
                            \fill[gray!25] (\x,\y) circle (0.05);
                        \fi
                    \fi
                \fi
            \fi
            
            \ifdim \x pt>0pt
                \ifdim \x pt<10pt
                    \ifdim \y pt>0pt
                        \ifdim \y pt<10pt
                            \ifdim \x pt>8pt
                                \fill[gray!85] (\x,\y) circle (0.05);
                            \fi
                            \ifdim \x pt<2pt
                                \fill[gray!85] (\x,\y) circle (0.05);
                            \fi
                            \ifdim \y pt>8pt
                                \fill[gray!85] (\x,\y) circle (0.05);
                            \fi
                            \ifdim \y pt<4pt
                                \fill[gray!85] (\x,\y) circle (0.05);
                            \fi
                        \fi
                    \fi
                \fi
            \fi
        }
        
        \foreach \i in {0,...,\gridsize} {
            \draw[gray!30, very thin] (0,\i) -- (\gridsize,\i);
            \draw[gray!30, very thin] (\i,0) -- (\i,\gridsize);
        }
        
        \draw[line width=0.8pt] (2,2) rectangle (8,8);

        \tikzstyle{spiral}=[->, >=stealth, shorten >=1pt]
        
        \foreach \i in {4,...,7} {
            \ifodd\i
                \draw[spiral] (2.5,\i+0.5) -- (7.5,\i+0.5);
            \else
                \draw[spiral] (7.5,\i+0.5) -- (2.5,\i+0.5);
            \fi
        }
        \foreach \i in {6,...,8} {
            \ifodd\i
                \draw[spiral] (2.5,\i-0.5) -- (2.5,\i-1.5);
            \else
                \draw[spiral] (7.5,\i-0.5) -- (7.5,\i-1.5);
            \fi
        }
        
        \node at (2.5,4.5) {\tiny UAV};
    
    \end{tikzpicture}
    
    }
\caption{Example of a boustrophedon path.}
\end{subfigure}\hfill
    \begin{subfigure}{0.3\linewidth}
    	
    	\resizebox{0.99\linewidth}{.99\linewidth}{
    		\begin{tikzpicture}
    		
    		\def\gridsize{10}
    		
    		\foreach \i in {1,...,1000} {
    			\pgfmathsetmacro{\x}{5 + (rand + rand + rand + rand + rand + rand + rand + rand + rand)}
    			\pgfmathsetmacro{\y}{5 + (rand + rand + rand + rand + rand + rand + rand + rand + rand)}
    			
    			\ifdim \x pt>0pt
	    			\ifdim \x pt<10pt
		    			\ifdim \y pt>0pt
			    			\ifdim \y pt<10pt
				    			\fill[gray!85] (\x,\y) circle (0.05);
			    			\fi
		    			\fi
	    			\fi
    			\fi
    			
    			\pgfmathtruncatemacro{\cellx}{floor(\x)}
    			\pgfmathtruncatemacro{\celly}{floor(\y)}
				
				\foreach \j in {4,...,7} {
    			\ifnum\cellx=\j
    				\ifnum\celly=4
		    			\fill[gray!25] (\x,\y) circle (0.05);
		    		\fi
    			\fi
	    		}
    	
			    \foreach \j in {3,...,5} {
				    \ifnum\cellx=\j
					    \ifnum\celly=5
						    \fill[gray!25] (\x,\y) circle (0.05);
					    \fi
				    \fi
				}
			
				\foreach \j in {3,...,5} {
					\ifnum\cellx=\j
						\ifnum\celly=6
							\fill[gray!25] (\x,\y) circle (0.05);
						\fi
					\fi
				}
			    
    		}
    		
    		\foreach \i in {0,...,\gridsize} {
    			\draw[gray!30, very thin] (0,\i) -- (\gridsize,\i);
    			\draw[gray!30, very thin] (\i,0) -- (\i,\gridsize);
    		}
    		
    		\draw[line width=0.8pt] (2,2) rectangle (8,8);

    		\tikzstyle{spiral}=[->, >=stealth, shorten >=1pt]
    		
    		\draw[spiral] (7.5,4.5) -- (5.5,4.5);
    		\draw[spiral] (5.5,4.5) -- (5.5,6.5);
    		\draw[spiral] (5.5,6.5) -- (3.5,6.5);
    		\draw[spiral] (3.5,6.5) -- (3.5,5.5);
    		\draw[spiral] (3.5,5.5) -- (4.5,5.5);
    		\draw[spiral] (4.5,5.5) -- (4.5,4.5);
    		
    		\node at (2.5,4.5) {\tiny UAV};
    		
    		\end{tikzpicture}
    	}
        \caption{Schema of a possible B\&B trajectory.}
    \end{subfigure}

    \caption{Schematic drawing of the three algorithms under investigation.}
    \label{fig:schematic_drawing}
    \vspace{-5mm}
\end{figure*}
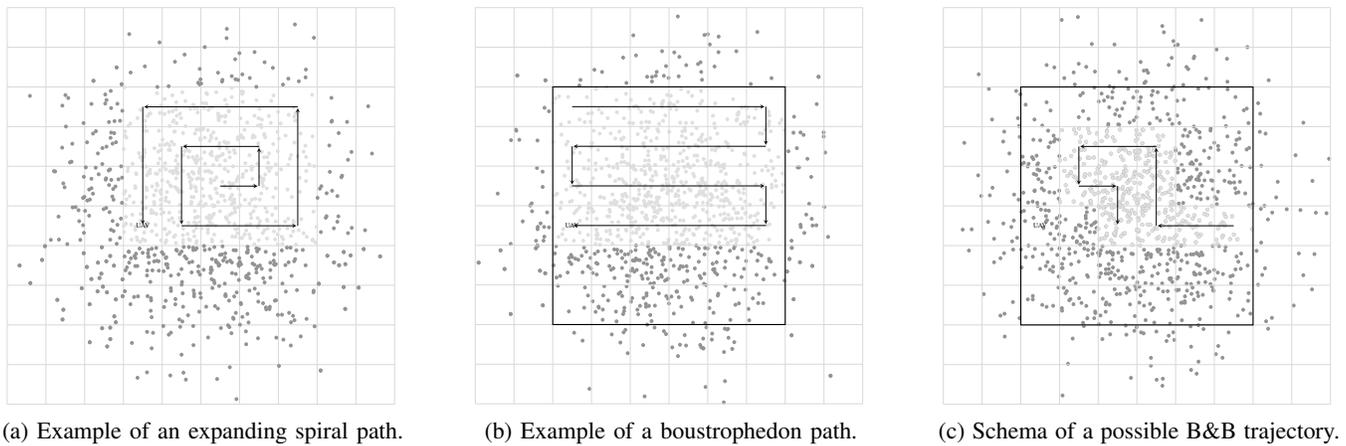

We compare distinct UAV path planning strategies specifically tailored for maritime search and rescue. Recognizing the dynamic nature of maritime environments, each strategy under consideration incorporates varying degrees of water current and wind information to enhance search efficiency. To provide a comprehensive evaluation, we looked into three strategies from the literature; the first two of them are used in practice \cite{costguardmanual} while the third \cite{sato2008path} aims to bridge the gap from theory to application. Fig. \ref{fig:schematic_drawing} shows a schematic drawing of their functionality.

\begin{enumerate}[wide=0pt]
    \item 
    Expanding Spiral Method: This strategy, already implemented by the Canadian and US Coast Guard \cite{costguardmanual}, begins its search from the presumed location of the search target. The UAV then follows an outward spiral pattern, progressively increasing the radius of the spiral as the search continues. The idea is to start the search close to the last known or estimated location and then expand outwards in a systematic manner. Furthermore, the estimated location is computed by using the particles, which incorporate water current and wind data to model their trajectory.
    
    \item 
    Boustrophedon Rectangle Planner: This approach entails the allocation of rectangular search zones across areas designated as high-probability zones. The UAV then conducts its search in a boustrophedon (back-and-forth) manner within each rectangle before moving to the next. Similar to the expanding spiral method, particles are used in tracking the estimated position of the search targets, adjusting for the influence of water current and wind dynamics over time. Again, this approach is already used in practice, see \cite{kratzke2010search}.

    \item 
    Global-Local Branch-and-Bound Method: Our proposed approach leverages a hybrid model for trajectory planning. On a global scale, similar to the boustrophedon method, this planner starts by identifying and estimating rectangular zones that encapsulate a predetermined portion of particles. The UAV first heads to these regions. Upon reaching the designated area, the approach shifts to a local scale. At this level, the UAV employs a modified version of the branch and bound algorithm \cite{sato2008path, clausen1999branch}, specifically leveraging the smaller search space. This two-tier system ensures the UAV covers vast areas quickly while maintaining the precision necessary for finer, more detailed searches. A more in-depth explanation of the details will be given later in this section.
\end{enumerate}

Fig. \ref{fig:schematic_drawing} illustrates the first two planning methods. Specifically, the spiral planner computes the center of gravity for each of the particle clouds. Then, it calculates the shortest path visiting all of them and moves from one to the next. Given that real-world situations typically have a small number of targets, one can comprehensively examine all possible permutations. Once arrived, it starts searching for a search target by performing an outward spiral until it observes a target. Next, it proceeds to the next search target. Over time, all search targets' estimated locations are updated according to the modeled probability distribution given by the particles.

The boustrophedon planner acts in a similar fashion by placing rectangular search areas on the map. For each particle cloud, it constructs a rectangle large enough to contain a predefined portion $\eta$ of each collection of particles. Later, in our experiments (section \ref{sec:experiments}), we chose $\eta = 0.75$, because this choice empirically gave the best results. Once arrived at a rectangle, the UAV traverses it in a boustrophedon fashion, meaning it is flying back and forth to cover the whole ground area in the rectangle.

\begin{algorithm}
\caption{Branch and Bound Planner}\label{alg:bnb}
\begin{algorithmic}
\Require Number of targets $n\in \mathbb N$
\Require Targets with associated particles $L = \{ (x_k, P_k) \}_{k=1}^n$
\Require Containment percentage $0<\eta <1$
\Require UAV position $u = (x,y)$
\State $L \gets \text{OrderedList}(L)$ \Comment{Order $L$ to form shortest path}
\State $R \gets \{ R_k  \}_{k=1}^n$ \Comment{Rectangles $R_k$ enclosing $\eta$ particles of $P_k$}
\For {$k \leq n; ~ k ++$}
\If{$u$ is at $R_k$}
    \State $u \gets$ \texttt{Branch\&BoundSearch}$(u, R_k)$
\Else
    \State $u \gets $\texttt{Advance}$(u, R_k)$
\EndIf
\State $R \gets $\texttt{UpdateSearchAreas}$(\{ R_k \})$
\EndFor
\end{algorithmic}
\end{algorithm}

Algorithm \ref{alg:bnb} shows the pseudo code for our branch and bound planner. The subprocedure '\texttt{Advance}' flys the UAV from its current position into the direction of the next rectangle in the queue. The call to '\texttt{UpdateSearchAreas}' recalculates the rectangular search areas to account for particle drift during the course of the simulation. Finally, '\texttt{Branch\&BoundSearch}' performs a branch and bound algorithm on the rectangular search area. For details of this algorithm see \cite{clausen1999branch}, for its specific application to trajectory planning problems see \cite{sato2008path}. The main differences in the use of branch and bound in this paper compared to \cite{sato2008path} is that we only apply it in designated search areas, reducing the problem space drastically. Furthermore, instead of computing a precise upper bound -- which is computationally expensive -- we employ a heuristic, accepting a possibly sub-optimal solution while making the algorithm applicable in practice. Specifically, the heuristic we employ merely computes the number of particles in a close vicinity to the investigated position -- locations with a higher number of particles nearby are valued higher to the algorithm than others. This is a greedy approach, yet it worked well in our experiments (section \ref{sec:experiments}).

\vspace{-1mm}
\subsection{Particle Filter with negative Measurements}
\label{sec:particlefilter}
Particle filters, commonly used for state estimation, rely on representing a system's uncertainty through a set of $N \in \mathbb N$ weighted samples (particles) that collectively describe the system's probabilistic belief over its state\textcolor{blue}{, usually denoted $\mathcal{M} = \{ (x_k, \omega_k) \vert 1 \leq k \leq N \}$ \cite{choset2005principles, elfring2021particle}}. An integral component of the particle filter process is the update step, where the weight \textcolor{blue}{$\omega_k$} of each particle is adjusted based on the likelihood of an observed measurement \textcolor{blue}{$y$} given that particle's state \textcolor{blue}{$x_k$, that is
\begin{align*}
\vspace{-4mm}
    \omega_k \leftarrow P(y \vert x_k) .
    \vspace{-4mm}
\end{align*}
}
The measurement process in this work's setup is special in the sense, that the UAV either observes a search target in a grid cell, or it does not. This translates to a particle filter in the following way. Assume our UAV is observing grid cell $(i,j)$ (denoted $\operatorname{cell}_{(i,j)}$) at time $t$. Then either we have $y = 1$, if the search target is contained in $\operatorname{cell}_{(i,j)}$, or $y=0$ otherwise. Hence, the relevant cases for the particle update are
\begin{align}
    \label{eq:delete}
    P \left( y = 0 \vert x_k \in \operatorname{cell}_{(i,j)} \right) & = \varepsilon, \\
    \label{eq:keep}
    P \left( y = 0 \vert x_k \not\in \operatorname{cell}_{(i,j)} \right) & = 1 - \varepsilon .
\end{align}
Here $0 \leq \varepsilon < 1$ accounts for sensor detection errors. Later, in our experiments (section \ref{sec:experiments}), we chose the flight altitude low enough that we can safely assume $\varepsilon = 0$.
Expression (\ref{eq:delete}) equals zero, because particle $x_k \in \operatorname{cell}_{(i,j)}$ at time $t$, yet we observed, that no search target is present. The weight update for particles which are not in $\operatorname{cell}_{(i,j)}$ is given by equation (\ref{eq:keep}). In the cases, where $y=1$, the UAV found the search target and we stop searching.
Therefore, in the resampling step of the particle filter, the particles contained in $\operatorname{cell}_{(i,j)}$ are resampled with probability $0$, thus being erased.

\vspace{-1mm}
\subsection{Search Targets' Movement Model}
\label{sec:opendrift}

\begin{figure*}[t]
	\centering
	\begin{subfigure}{0.32\textwidth}
            \includegraphics[width=.99\linewidth]{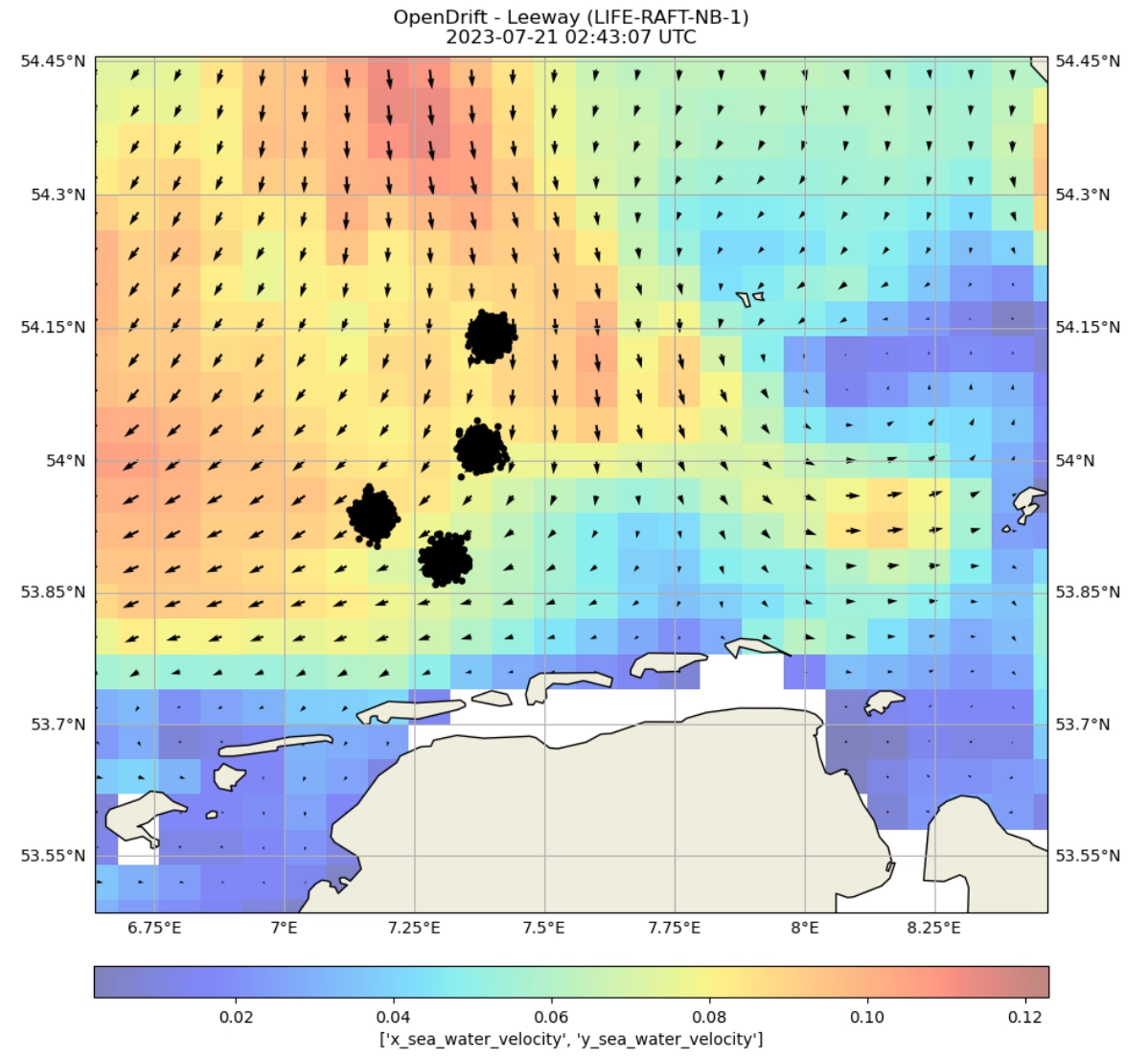}
            \caption{At the start of the simulation.}
	\end{subfigure}
        \begin{subfigure}{0.32\textwidth}
            \includegraphics[width=.99\linewidth]{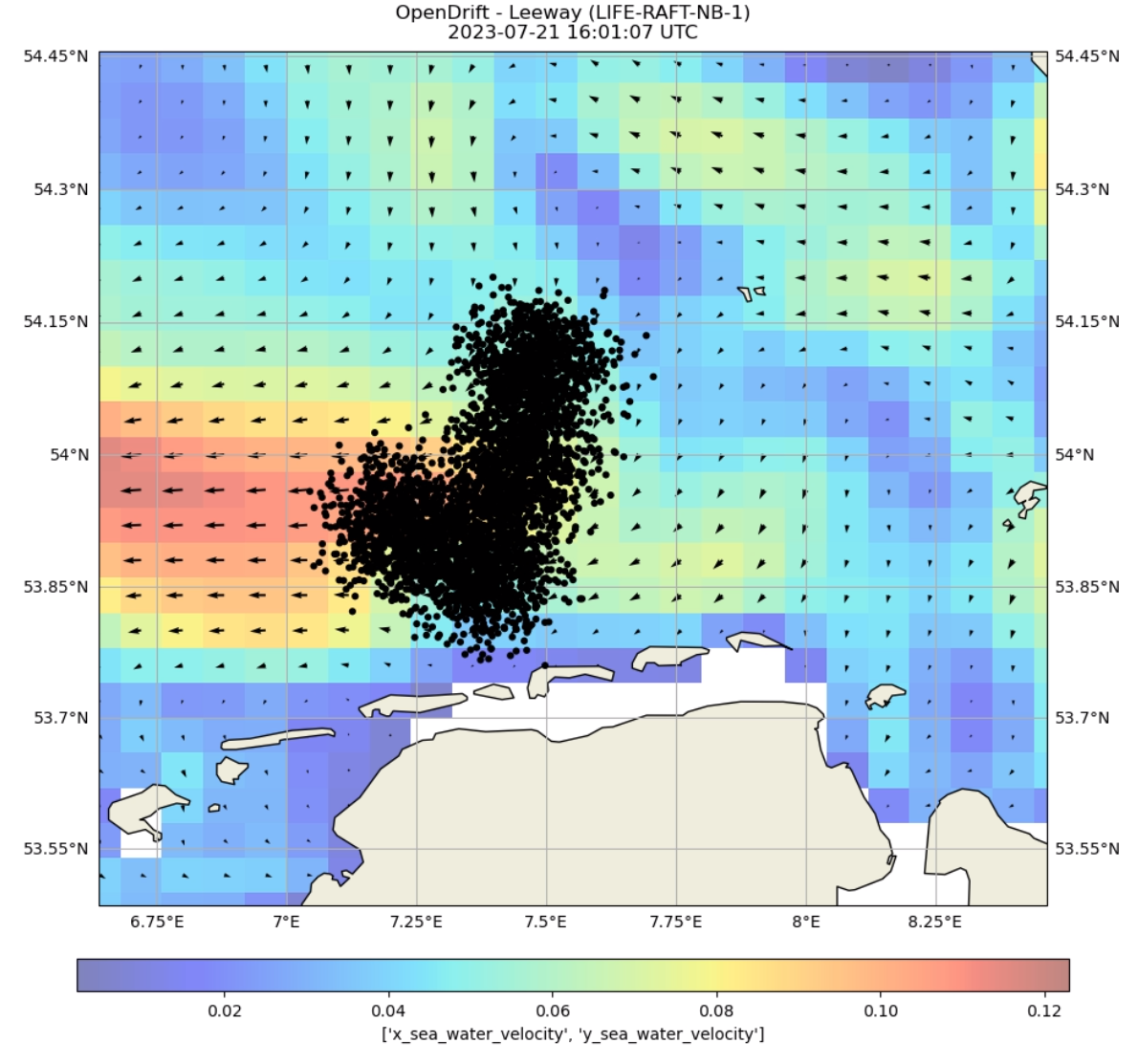}
            \caption{After twelve hours of the simulation.}
        \end{subfigure}
            \begin{subfigure}{0.32\textwidth}
    	\includegraphics[width=.99\linewidth]{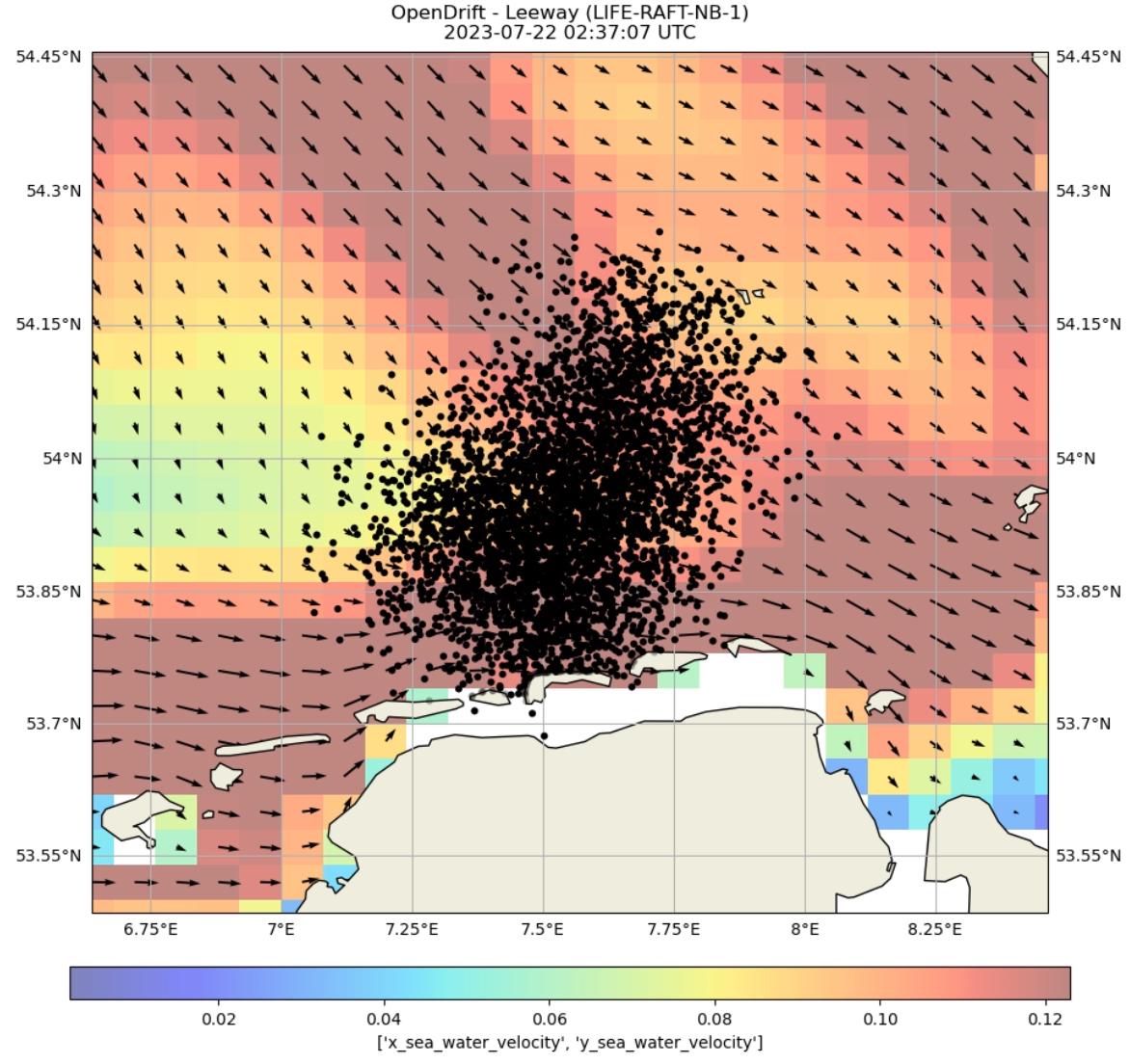}
	\caption{After $24$ hours of the simulation.}
        \end{subfigure}
    \caption{Three output plots from the OpenDrift framework. It simulates four search targets, observable as four particle clouds in the left image, showing the start of the simulation. The middle and right images show the simulation after twelve and $24$ hours. The background shows the underlying water flow, changing over time. The location is roughly at $54.0 ~ \text{N}, 7.5 ~ \text{E}$.}
    \label{fig:opendrift}
\end{figure*}

In examining the dynamics of distressed search targets and their movement patterns, this study utilizes the OpenDrift framework \cite{dagestad2018opendrift}. This open-source tool offers a reliable simulation platform for a variety of floating objects, like ships and life rafts, while also including debris and oil spills in aquatic settings. Notably, OpenDrift's design draws from the simulation models employed by the United States Coast Guards' planning software \cite{kratzke2010search}. The following provides a brief summary of the elements relevant to this research. Figure \ref{fig:opendrift} shows examples from an OpenDrift simulation.
In maritime search and rescue operations, accurately predicting the movement of search targets in the ocean is of high importance. At the beginning of a search mission, represented as $t = 0 $, a total of $ n \in \mathbb N$ particles are sampled from a bivariate Gaussian distribution surrounding the initial belief for the position of each search target. The variance of this distribution corresponds to the inherent uncertainty in our initial beliefs regarding the precise location of the targets.
If the uncertainty is smaller or larger, so is the variance from which we sample the particles.
To model the trajectory of these particles over subsequent time intervals, we employ the Lagrangian movement model \cite{breivik2008operational} embedded within the OpenDrift framework. This model offers a robust estimation of each particle's path, taking into account the influences of both water currents and wind flow. To more comprehensively mirror the real-world scenario, OpenDrift incorporates a minor diffusion factor into the trajectory of each particle. This factor serves to simulate the progressively increasing uncertainty associated with the position of the search targets over time.
The wind and water flow information is gathered from \cite{hycom} and \cite{iwamoto2016ocean}, respectively. Their respective provided resolutions are roughly $5 ~km^2$ and $67 \times 33 ~ km^2$.

\section{Experiments}
\label{sec:experiments}

The main goal of this chapter is to emulate real-world conditions as closely as possible. We aim to bridge the significant gap between algorithms, which are good in theory but not applicable, and their tangible, practical applications in maritime SAR operations. The following experiments reflect that as well.

One of the critical factors we had to consider was the grid tile's size. For our simulations to reflect real-life conditions, it is essential to capture the realistic surface area that a standard optical sensor would cover when deployed on a UAV.
Previous work shows \cite{varga2022seadronessee, kiefer2022leveraging}, that any human in the water should be detected by an object detector running on the UAV when flying at around $100 ~ m$ altitude. Therefore, we've determined that a square tile measuring approximately $100 ~ m$ on each side represents the area a UAV would typically survey when flying at this altitude. That ensures, that we can safely mark a particle as observed once found in the same grid tile as the UAV.

Furthermore, the UAV's speed is based on the capabilites of smaller fixed-wing drones. With a modeled speed set at $18 ~ \nicefrac{m}{s}$ (\cite{elevonx_datasheet, trinity_datasheet}), our simulations mirror the typical operational speeds of these drones, which we argue are the most fitting for maritime SAR mission -- they are affordable, relatively easy to use, and compromise in between the flexibility of multi copters and the endurance and speed of small air crafts while not requiring a pilot \cite{messmer2024uav}.

\begin{table}[h]
	\centering
	\begin{tabular}{c|c|c|c}
		air speed & altitude & field of view & flight time \\\hline
		$18 ~ \nicefrac{m}{s}$ & $100 ~ m$ & $100 \times 100 ~ m^2$ & $1 - 5 ~ h$
	\end{tabular}
	\vspace*{5mm}
	\caption{Specifications of simulated UAV.}
	\label{tab:uav_specs}
\end{table}

Smaller fixed-wing drones equipped with electrical engines have approximately $1 - 2$ hours of average battery life span (\cite{elevonx_datasheet, trinity_datasheet}) while employing a combustion engine this class of UAVs can reach a flight time of $5$ hours (\cite{elevonx_datasheet}). In practice, electrical engines are less error-prone and require less maintenance. Also, in case of an accident, the environmental burden is smaller compared to engines running on gasoline.

To cover both cases in our simulations, we conducted experiments with a battery life span of $1,2,$ and $5$ hours, trying to be as realistic and close to practical application as possible. Table \ref{tab:uav_specs} shows an overview of the technical specifications of the simulated UAV. \\
In all described experiments contained in this section, we used $10,000$ particles per search target for the OpenDrift simulation. They were sampled roughly $2 ~ km$ around the simulated position of each target. The grid size used is $250 \times 250 ~ km^2$ with a tile size of $100 ~m$, resulting in a $2500 \times 2500$ grid. \newline 
In all experiments, the UAV's take-off area is right on shore. Precisely, it takes off at $53.722827 ~N, 7.192965 ~ E$. Search targets were sampled at a distance of $10,20,$ or $30$ kilometers in the sea. The experiments become more challenging for the agents as the distance increases because the drone requires more time to reach the search area. This delay allows the particles to disperse more before the UAV begins its search. For example, at a speed of $18 ~ \nicefrac{m}{s}$ (see Table \ref{tab:uav_specs}), the UAV needs roughly $9.5$ minutes to fly $10$ kilometers, but almost $28$ minutes to travel $30$ kilometers.
In all experiments, there are either one or two targets. First, we randomly drew whether to sample one or two targets based on a uniform distribution. Next, we sampled the specified number of search targets at the respective distance.
We report the success rates, the time to detect the first search target, and the success rate for finding the first search target only. These metrics are calculated as the average across all runs with identical distances to the targets. For each table and agent, the reported numbers are averages taken over roughly 500 runs.

Of course the most important metric is the success rate, defined as the number of found targets divided by the total number of search targets that could have been found. The other two additional metrics provide further valuable insights. Although secondary, it is also crucial to optimize for finding search targets quickly; for example, if the targets are humans, a quicker detection helps prevent them from cooling down too much.
The success rate for only the first search targets, when compared with the overall success rate, shows whether the agent was able to detect the second target as well. This comparison is valuable because it reveals how the agents handle higher uncertainty, as these targets have been adrift for a longer period, causing their corresponding particles to disperse more. This is similar to the scenarios on search targets at larger distances. Roughly equal values for 'success rate' and 'success rate $1^\text{st}$' suggest, that the agent was generally successful in finding the second search target. On the other hand, if there is a significant difference between the two -- relative to the overall success rate -- it indicates that the second search target was often missed.

The tables are arranged in ascending order based on the distance from the UAV's take-off to the search targets. The lines 'Spiral' and 'Boustrophedon' show the results for the respective agents as described in section \ref{sec:methods}. In these experiment, for the boustrophedon agent the portion of particles that is contained in a rectangle to be searched is $0.75$. 'B\&B 15', 'B\&B 35', and 'B\&B 50' denote the branch and bound planners from section \ref{sec:methods}; the number indicates the maximum duration, in minutes, that the UAV dedicates to searching for a target inside a rectangle using the branch and bound method. Specifically, the first seeks for $15$ minutes, the second for up to $35$ minutes, and the last for $50$ minutes. Once the UAV identifies a search target, it immediately halts the search and moves on to the next target, indifferent to the time already spent.

%
%

\begin{table}[]
	\begin{center}
		
		\begin{tabular}{cccc}
			& success rate $\uparrow$ & time $1^\text{st}$ $\downarrow$ & success rate $1^\text{st}$  $\uparrow$ \\
			\cmidrule(l){2-4} 
			\multicolumn{1}{c|}{Spiral}        &    $\mathbf{0.73}$   &    $15.0$ min.   &  $\mathbf{0.80}$  \\
			\multicolumn{1}{c|}{Boustrophedon} &   $0.48$    &   $46.1$ min.   &  $0.60$ \\
			\multicolumn{1}{c|}{B\&B 15}  &   $0.51$    &    $21.0$ min.   &   $0.65$   \\
			\multicolumn{1}{c|}{B\&B 35} &   $0.52$    &    $23.0$ min.   &  $0.67$  \\
			\multicolumn{1}{c|}{B\&B 50} &   $0.62$    &    $26.9$ min.   &  $0.72$  \\
		\end{tabular}
		
	\end{center}
	\caption{Experimental results for search targets roughly $10 ~ km$ off shore. The 'time $1^\text{st}$' values represent the average number of minutes for the respective agent to locate the first search target. The values for 'success rate $1^\text{st}$' indicate the success rate for finding the first search target only. Each line shows an average over roughly 500 runs.}
	\label{tab:_results_10km}
\end{table}

Table \ref{tab:_results_10km} shows the results for the experiments where the search targets were sampled roughly $10 ~ km$ away from the UAV's take off. Judging by their performance, it is the easiest task for the planning algorithms. This makes sense, as the UAV does not need to fly far before arriving at the position of the first targets. Hence the uncertainty about the position while searching is relatively low. Especially the spiral agent profits from that, as it starts its search at the point of highest probability, the center of the particle cloud, making it the best performing planner in this case. Its time of success for the first target is close to optimal.
We were surprised by the boustrophedon planner's worse performance compared to the spiral. We argue that is due to the rectangular agent starting its search at the edge of the particle cloud, a low probability area. This allows the particles to disperse before exploring areas with a higher likelihood.
The branch-and-bound agents, B\&B$15$, B\&B$35$, and B\&B$50$ perform well, each outperforming the boustrophedon planner in this scenario. However, they are unable to unfold their full potential in the easiest scenario. This can be seen by the most capable one, B\&B$50$, still trailing the spiral agent by roughly $10$ percentage points.
In general, the success rates for the first target are quite close to the overall success rates, indicating that all agents are generally capable of finding the second target as well.

%
%

\begin{table}[]
	\begin{center}
		
		\begin{tabular}{@{}cccc@{}}
			& success rate $\uparrow$ & time $1^\text{st}$ $\downarrow$ & success rate $1^\text{st}$  $\uparrow$ \\
			\cmidrule(l){2-4} 
			\multicolumn{1}{c|}{Spiral}        &   $0.44$    &   $30.2$ min.  &   $0.55$ \\
			\multicolumn{1}{c|}{Boustrophedon} &   $0.43$    &   $54.2$ min.   & $0.57$  \\
			\multicolumn{1}{c|}{B\&B 15}  &   $0.35$    &   $32.2$ min.  &   $0.46$ \\
			\multicolumn{1}{c|}{B\&B 35} &    $0.44$   &    $35.9$ min.  &  $0.60$  \\
			\multicolumn{1}{c|}{B\&B 50} &    $\mathbf{0.54}$   &    $41.3$ min.  & $\mathbf{0.68}$ \\
		\end{tabular}
		
	\end{center}
	\caption{Experimental results for search targets roughly $20 ~ km$ off shore. The 'time $1^\text{st}$' values represent the average number of minutes for the respective agent to locate the first search target. The values for 'success rate $1^\text{st}$' indicate the success rate for finding the first search target only. Each line shows an average over roughly 500 runs.}
	\label{tab:_results_20km}
\end{table}

Table \ref{tab:_results_20km} shows the search results for targets sampled at a distance of roughly $20 ~km$. In these experiments, 
we see the spiral agents' performance deteriorating quickly (compared to Table \ref{tab:_results_10km}) as the uncertainty of the search targets' location grows.
It starts searching in the center of a particle cloud, but since this cloud's particles already dispersed by the time the UAV arrives, the search target does not need to be near the center. Therefore, in some cases, the UAV is simply not fast enough to reach the search target because the drone works its way outward in increasingly larger circles. However, in the successful cases, this agent is the quickest at finding the search target.
Surprisingly, the boustrophedon agent's performance does not decrease as much. However, the time spent on finding the first search target is significantly higher than the others, indicating that it generally locates targets through persistence rather than efficiency.
For the branch and bound agents, while their performance also decreases, it is not as drastic as for the spiral agent. Also, they deliver the best search performance for this scenario.
The ratio between success rate for the first target and overall success rate\footnote{For each run, we sampled either one or two search targets with uniform probability. Therefore, there are approximately as many runs with two targets as there are with one target or, roughly speaking, twice as many first search targets as there are second search targets over all runs. Hence, for an agent never finding the second search target, we would expect a ratio of $\nicefrac{3}{2}$ between 'success rate $1^\text{st}$' and 'success rate'. Conversely, an agent with comparable performance in finding both the first and the second search targets would have a ratio close to $1$ between these two quantities.} shows, that the spiral agent and B\&B$50$ are the agents with the best balance for the trade-off between searching for the first target and giving up on it in favour of a chance on finding the second.


%
%

\begin{table}[]
	\begin{center}
		
		\begin{tabular}{@{}cccc@{}}
			& success rate $\uparrow$ & time $1^\text{st}$ $\downarrow$ & success rate $1^\text{st}$  $\uparrow$ \\
			\cmidrule(l){2-4} 
			\multicolumn{1}{c|}{Spiral}        &    $0.10$   &   $59.0$ min.   &   $0.14$  \\
			\multicolumn{1}{c|}{Boustrophedon} &   $0.17$    &    $73.8$ min.    &  $0.23$  \\
			\multicolumn{1}{c|}{B\&B 15}  &    $0.16$   &   $51.2$ min.  &  $0.23$  \\
			\multicolumn{1}{c|}{B\&B 35} &    $0.18$   &    $51.3$ min.  &  $0.27$  \\
			\multicolumn{1}{c|}{B\&B 50} &    $\mathbf{0.30}$   &    $58.5$ min.   &    $\mathbf{0.37}$  \\
		\end{tabular}
		
	\end{center}
	\caption{Experimental results for search targets roughly $30 ~ km$ off shore. The 'time $1^\text{st}$' values represent the average number of minutes for the respective agent to locate the first search target. The values for 'success rate $1^\text{st}$' indicate the success rate for finding the first search target only. Each line shows an average over roughly 500 runs.}
	\label{tab:_results_30km}
\end{table}

Table \ref{tab:_results_30km} shows the results for experiments with targets at a distance of roughly $30~km$ from the UAV's take-off position. 
Notably, the performance of all agents is relatively low compared to the results shown in Tables \ref{tab:_results_10km} and \ref{tab:_results_20km}, confirming the assumption, that this is the hardest task, as the search targets are sampled at the largest distance to the take-off position.
We observe that B\&B50, the agent investing the most resources in finding each target, achieves the highest performance, both overall and for the first search target. 
The ratio between success rate for the first target and overall success rate is the lowest for B\&B50, meaning that it is the agent most successful at finding the second search targets${}^3$. That is, although it spends quite some time on the first search target, making the agent's belief about the second target's location very uncertain.
Notably, the spiral agent's search time is not the smallest for this task. 

\section{Conclusion and Outlook}

We have developed an improved path planning algorithm for UAVs in maritime search and rescue missions. Recognizing the challenges of dynamic maritime conditions and uncertain target locations, our method integrates real-time meteorological data and probabilistic models. This offers a more adaptive and effective approach than existing solutions.

Our research also highlighted the subtle differences of traditional particle filters when primarily faced with negative measurements in maritime contexts.

Looking forward, we aim to further refine our algorithms considering the interplay of environmental dynamics and target motion. Investigating coordinated multi-UAV missions could also enhance search and rescue operations. Ultimately, our goal remains to transition these theoretical advancements into practical applications.







\bibliographystyle{IEEEtran}
\bibliography{root}

\begin{thebibliography}{10}
\providecommand{\url}[1]{#1}
\csname url@rmstyle\endcsname
\providecommand{\newblock}{\relax}
\providecommand{\bibinfo}[2]{#2}
\providecommand\BIBentrySTDinterwordspacing{\spaceskip=0pt\relax}
\providecommand\BIBentryALTinterwordstretchfactor{4}
\providecommand\BIBentryALTinterwordspacing{\spaceskip=\fontdimen2\font plus
\BIBentryALTinterwordstretchfactor\fontdimen3\font minus
  \fontdimen4\font\relax}
\providecommand\BIBforeignlanguage[2]{{%
\expandafter\ifx\csname l@#1\endcsname\relax
\typeout{** WARNING: IEEEtran.bst: No hyphenation pattern has been}%
\typeout{** loaded for the language `#1'. Using the pattern for}%
\typeout{** the default language instead.}%
\else
\language=\csname l@#1\endcsname
\fi
#2}}

\bibitem{russell2010artificial}
S.~J. Russell, \emph{Artificial intelligence a modern approach}.\hskip 1em plus
  0.5em minus 0.4em\relax Pearson Education, Inc., 2010.

\bibitem{varga2022seadronessee}
L.~A. Varga, B.~Kiefer, M.~Messmer, and A.~Zell, ``Seadronessee: A maritime
  benchmark for detecting humans in open water,'' in \emph{Proceedings of the
  IEEE/CVF winter conference on applications of computer vision}, 2022, pp.
  2260--2270.

\bibitem{lee2018snip}
N.~Lee, T.~Ajanthan, and P.~H. Torr, ``Snip: Single-shot network pruning based
  on connection sensitivity,'' \emph{arXiv preprint arXiv:1810.02340}, 2018.

\bibitem{varga2021tackling}
L.~A. Varga and A.~Zell, ``Tackling the background bias in sparse object
  detection via cropped windows,'' in \emph{Proceedings of the IEEE/CVF
  International Conference on Computer Vision}, 2021, pp. 2768--2777.

\bibitem{messmer2022gaining}
M.~Messmer, B.~Kiefer, and A.~Zell, ``Gaining scale invariance in uav bird’s
  eye view object detection by adaptive resizing,'' in \emph{2022 26th
  International Conference on Pattern Recognition (ICPR)}.\hskip 1em plus 0.5em
  minus 0.4em\relax IEEE, 2022, pp. 3588--3594.

\bibitem{kiefer2022leveraging}
B.~Kiefer, D.~Ott, and A.~Zell, ``Leveraging synthetic data in object detection
  on unmanned aerial vehicles,'' in \emph{2022 26th International Conference on
  Pattern Recognition (ICPR)}.\hskip 1em plus 0.5em minus 0.4em\relax IEEE,
  2022, pp. 3564--3571.

\bibitem{raap2019moving}
M.~Raap, M.~Preu{\ss}, and S.~Meyer-Nieberg, ``Moving target search
  optimization--a literature review,'' \emph{Computers \& Operations Research},
  vol. 105, pp. 132--140, 2019.

\bibitem{sato2008path}
H.~Sato, ``Path optimization for single and multiple searchers: models and
  algorithms,'' Ph.D. dissertation, Citeseer, 2008.

\bibitem{raap2017trajectory}
M.~Raap, M.~Zsifkovits, and S.~Pickl, ``Trajectory optimization under
  kinematical constraints for moving target search,'' \emph{Computers \&
  Operations Research}, vol.~88, pp. 324--331, 2017.

\bibitem{martinez2021search}
I.~Martinez-Alpiste, G.~Golcarenarenji, Q.~Wang, and J.~M. Alcaraz-Calero,
  ``Search and rescue operation using uavs: A case study,'' \emph{Expert
  Systems with Applications}, vol. 178, p. 114937, 2021.

\bibitem{du2019visdrone}
D.~Du, P.~Zhu, L.~Wen, X.~Bian, H.~Lin, Q.~Hu, T.~Peng, J.~Zheng, X.~Wang,
  Y.~Zhang, \emph{et~al.}, ``Visdrone-det2019: The vision meets drone object
  detection in image challenge results,'' in \emph{Proceedings of the IEEE/CVF
  international conference on computer vision workshops}, 2019, pp. 0--0.

\bibitem{morin2010ant}
M.~Morin, L.~Lamontagne, I.~Abi-Zeid, and P.~Maupin, ``The ant search
  algorithm: An ant colony optimization algorithm for the optimal searcher path
  problem with visibility,'' in \emph{Advances in Artificial Intelligence: 23rd
  Canadian Conference on Artificial Intelligence, Canadian AI 2010, Ottawa,
  Canada, May 31--June 2, 2010. Proceedings 23}.\hskip 1em plus 0.5em minus
  0.4em\relax Springer, 2010, pp. 196--207.

\bibitem{berger2013exact}
J.~Berger, N.~Lo, and M.~Noel, ``Exact solution for search-and-rescue path
  planning,'' \emph{International Journal of Computer and Communication
  Engineering}, vol.~2, no.~3, p. 266, 2013.

\bibitem{riehl2007cooperative}
J.~R. Riehl, G.~E. Collins, and J.~P. Hespanha, ``Cooperative graph-based model
  predictive search,'' in \emph{2007 46th IEEE Conference on Decision and
  Control}.\hskip 1em plus 0.5em minus 0.4em\relax IEEE, 2007, pp. 2998--3004.

\bibitem{dagestad2018opendrift}
K.-F. Dagestad, J.~R{\"o}hrs, {\O}.~Breivik, and B.~{\AA}dlandsvik, ``Opendrift
  v1. 0: a generic framework for trajectory modelling,'' \emph{Geoscientific
  Model Development}, vol.~11, no.~4, pp. 1405--1420, 2018.

\bibitem{wu2023modeling}
J.~Wu, L.~Cheng, and S.~Chu, ``Modeling the leeway drift characteristics of
  persons-in-water at a sea-area scale in the seas of china,'' \emph{Ocean
  engineering}, vol. 270, p. 113444, 2023.

\bibitem{guoxiang2010sargis}
L.~Guoxiang and L.~Maofeng, ``Sargis: A gis-based decision-making support
  system for maritime search and rescue,'' in \emph{2010 International
  Conference on E-Business and E-Government}.\hskip 1em plus 0.5em minus
  0.4em\relax IEEE, 2010, pp. 1571--1574.

\bibitem{kratzke2010search}
T.~M. Kratzke, L.~D. Stone, and J.~R. Frost, ``Search and rescue optimal
  planning system,'' in \emph{2010 13th International Conference on Information
  Fusion}.\hskip 1em plus 0.5em minus 0.4em\relax IEEE, 2010, pp. 1--8.

\bibitem{li2023survey}
J.~Li, G.~Zhang, C.~Jiang, and W.~Zhang, ``A survey of maritime unmanned search
  system: Theory, applications and future directions,'' \emph{Ocean
  Engineering}, vol. 285, p. 115359, 2023.

\bibitem{tiemann2018supporting}
J.~Tiemann, O.~Feldmeier, and C.~Wietfeld, ``Supporting maritime search and
  rescue missions through uas-based wireless localization,'' in \emph{2018 IEEE
  Globecom Workshops (GC Wkshps)}.\hskip 1em plus 0.5em minus 0.4em\relax IEEE,
  2018, pp. 1--6.

\bibitem{costguardmanual}
``{Canadian Coast Guard Auxiliary Search \& Rescue Crew Manual},''
  \url{https://ccga-pacific.org/files/library/Chapter_9_Search.pdf}, accessed:
  2023-08-16.

\bibitem{clausen1999branch}
J.~Clausen, ``Branch and bound algorithms-principles and examples,''
  \emph{Department of Computer Science, University of Copenhagen}, pp. 1--30,
  1999.

\bibitem{choset2005principles}
H.~Choset, K.~M. Lynch, S.~Hutchinson, G.~A. Kantor, and W.~Burgard,
  \emph{Principles of robot motion: theory, algorithms, and
  implementations}.\hskip 1em plus 0.5em minus 0.4em\relax MIT press, 2005.

\bibitem{elfring2021particle}
J.~Elfring, E.~Torta, and R.~van~de Molengraft, ``Particle filters: A hands-on
  tutorial,'' \emph{Sensors}, vol.~21, no.~2, p. 438, 2021.

\bibitem{breivik2008operational}
{\O}.~Breivik and A.~A. Allen, ``An operational search and rescue model for the
  norwegian sea and the north sea,'' \emph{Journal of Marine Systems}, vol.~69,
  no. 1-2, pp. 99--113, 2008.

\bibitem{hycom}
``{HYbrid Coordinate Ocean Model (HYCOM)},'' \url{https://www.hycom.org/},
  accessed: 2023-08-29.

\bibitem{iwamoto2016ocean}
M.~M. Iwamoto, F.~Langenberger, and C.~E. Ostrander, ``Ocean observing: serving
  stakeholders in the pacific islands,'' \emph{Marine Technology Society
  Journal}, vol.~50, no.~3, pp. 47--54, 2016.

\bibitem{elevonx_datasheet}
``{Data Sheet ElevonX SkyEye},''
  \url{https://www.elevonx.com/wp-content/uploads/2022/10/ElevonX.pdf},
  accessed: 2023-08-17.

\bibitem{trinity_datasheet}
``{Data Sheet Trinity F90+},''
  \url{https://quantum-systems.com/wp-content/uploads/2023/01/QS_TrinityF90_Overview_220912.pdf},
  accessed: 2023-08-20.

\bibitem{messmer2024uav}
M.~Messmer, B.~Kiefer, L.~A. Varga, and A.~Zell, ``Uav-assisted maritime search
  and rescue: A holistic approach,'' in \emph{2024 International Conference on
  Unmanned Aircraft Systems (ICUAS)}.\hskip 1em plus 0.5em minus 0.4em\relax
  IEEE, 2024, pp. 272--280.

\end{thebibliography}

\addtolength{\textheight}{-12cm}   

\end{document}